\documentclass[nohyperref]{article}

\usepackage{microtype}
\usepackage{graphicx}
\usepackage{subfigure}
\usepackage{booktabs}

\usepackage{hyperref}

\usepackage[accepted]{icml2022}

\usepackage{amsmath}
\usepackage{amssymb}
\usepackage{mathtools}
\usepackage{amsthm}

\usepackage[capitalize,noabbrev]{cleveref}

\theoremstyle{plain}

\theoremstyle{definition}

\theoremstyle{remark}

\usepackage[textsize=tiny]{todonotes}
\usepackage{tabularx, multirow,makecell}

\newcommand{\xlsrp}{XLS-R }
\newcommand{\tbh}[1]{\textbf{#1}}

\icmltitlerunning{Self-supervised Learning with Random-projection Quantizer for Speech Recognition}

\begin{document}

\twocolumn[
\icmltitle{Self-Supervised Learning with Random-Projection Quantizer \\for Speech Recognition}

\icmlsetsymbol{equal}{*}

\begin{icmlauthorlist}
\icmlauthor{Chung-Cheng Chiu}{equal,google}
\icmlauthor{James Qin}{equal,google}
\icmlauthor{Yu Zhang}{google}
\icmlauthor{Jiahui Yu}{google}
\icmlauthor{Yonghui Wu}{google}
\end{icmlauthorlist}

\icmlaffiliation{google}{Google Research, Brain Team}

\icmlcorrespondingauthor{Chung-Cheng Chiu}{chungchengc@google.com}
\icmlcorrespondingauthor{James Qin}{jamesqin@google.com}
\icmlcorrespondingauthor{Yu Zhang}{ngyuzh@google.com}

\icmlkeywords{self-supervised learning, speech recognition}

\vskip 0.3in
]

\printAffiliationsAndNotice{\icmlEqualContribution}

\begin{abstract}

We present a simple and effective self-supervised learning approach for speech recognition. The approach learns a model to predict the masked speech signals, in the form of discrete labels generated with a random-projection quantizer. In particular the quantizer projects speech inputs with a randomly initialized matrix, and does a nearest-neighbor lookup in a randomly-initialized codebook. Neither the matrix nor the codebook is updated during self-supervised learning. Since the random-projection quantizer is not trained and is separated from the speech recognition model, the design makes the approach flexible and is compatible with universal speech recognition architecture. On LibriSpeech our approach achieves similar word-error-rates as previous work using self-supervised learning with non-streaming models, and provides lower word-error-rates and latency than wav2vec 2.0 and w2v-BERT with streaming models. On multilingual tasks the approach also provides significant improvement over wav2vec 2.0 and w2v-BERT.
\end{abstract}

\section{Introduction}\label{intro}

Self-supervised learning has shown impressive improvement for the quality of the speech recognition models in recent years~\cite{wav2vec,vq-wav2vec,discretebert,wav2vec2,hubert,ssllimit,w2vbert,bigssl}. These learning approach enable the model to learn from unsupervised data and combine with supervised learning to improve the recognition accuracy. The capability of learning from unsupervised data is particularly beneficial when the supervised data is limited and opens up new opportunities for low resource languages and domains.

One common design principle of self-supervised learning for speech recognition centers around learning representations. Inspired by the success of BERT~\cite{bert}, one research trend in the speech community is to build BERT-inspired algorithms. One challenge in building BERT-style self-supervised learning for speech is to bridge the gap between continuous speech signals and the discrete text tokens, and a solution for addressing this issue is through learning speech representation~\cite{wav2vec,wav2vec2} or learning quantized representation~\cite{wav2vec2,vq-wav2vec,discretebert,hubert,w2vbert}. Many previous works proposed effective algorithms for learning speech representations, and the quantized result of those learned representations showed encouraging correlation with the phoneme of the utterances.

While representation learning is a critical topic for the speech field, combining it with self-supervised learning leads to two limitations that can slow the research progress: (1) Model architecture limitation. The integration of representation learning and self-supervised learning often requires the model to act the role of providing speech representation while still being effective for the downstream tasks. An effective representation model, however, may not always be effective for the downstream tasks. For example, a good representation learning model may require accessing the future context of the utterance, while downstream tasks may require a low latency model which prohibits the access of the future context. (2) Increased complexity. The objectives of representation learning and self-supervised learning are not always aligned, and the complexity of designing both algorithms and finding their balance can impede the research development. This complexity can also motivate the field toward designing more complicated algorithms instead of finding a simple and effective alternative.

In this work we propose \tbh{BE}RT-based \tbh{S}peech pre-\tbh{T}raining with \tbh{R}andom-projection \tbh{Q}uantizer (BEST-RQ), a simple and effective self-supervised learning algorithm for speech recognition. The algorithm masks speech signals and feeds them to the encoder part of the speech recognition model, and the encoder learns to predict the masked region based on the unmasked speech signals where the learning targets are labels provided by a random-projection quantizer. The random projection quantizer projects speech signals to a randomly initialized matrix, and finds a nearest vector in a randomly initialized codebook. The index of that vector is the target label. Neither the projection matrix nor the codebook is updated throughout the learning process. The quantizer does not require representation learning, and its separation from the model removes the limitation on the architecture design of the model. Despite its simplicity, on LibriSpeech the algorithm achieves similar results as previous work with non-streaming models, and provides better improvement with streaming models  compared with previous approaches. On multilingual tasks, the algorithm exhibits further gains compared to wav2vec 2.0~\cite{wav2vec2} and w2v-BERT~\cite{w2vbert}.

We conduct further analysis on the relation between representation learning quality and the self-supervised learning quality, and demonstrate that the two objectives are not inherently aligned in \cref{sec:vq-vae} and \cref{sec:scaling}. Such an observation is central to our design of self-supervised learning without representation learning, and opens up a new, less complicated research direction for self-supervised learning.

\section{Related Work}\label{related}

Many of the previous work on self-supervised learning for speech recognition focus on learning speech representation. wav2vec~\cite{wav2vec} applies contrastive learning to learn the future representation based on the past context. vq-wav2vec~\cite{vq-wav2vec} uses wav2vec to learn the representations and quantizes them to discrete tokens, and performs BERT-style pre-training to further improve the representation learning. DiscreteBERT~\cite{discretebert} extends vq-wav2vec by fine-tuning the BERT-pre-trained model on the downstream tasks. wav2vec 2.0~\cite{wav2vec2} uses contrastive learning with both past and future context to predict the representation of the masked parts. HuBERT~\cite{hubert} uses k-means to learn the initial quantizer that maps speech signals to discrete labels, and performs BERT-style pre-training where the inputs are masked speech signals and prediction targets are discrete labels. HuBERT further uses the pre-trained model as the new quantizer to train a new iteration of the model, and repeat the process to iteratively improve the pre-training results. w2v-BERT~\cite{w2vbert} uses a sub-network of the model to perform contrastive learning to learn speech representation, and use the rest of the network to perform BERT-style pre-training. w2v-BERT trains the representation learning and the BERT-style pre-training simultaneously. Our approach distinguishes from these work in avoiding the requirement of representation learning and separating the quantizer from the speech recognition model.

Our quantizer project input signals with a random matrix, which is similar to performing dimension reduction for the input signals. Using such quantization results as prediction target for self-supervised learning share a similar structure as the masked autoencoder (MAE)~\cite{mae}, which directly reconstruct the masked input signals. Another similar work in the computer vision community is BEiT~\cite{beit}, which trains a VQ-VAE~\cite{oord2018neural} as the quantizer and use the VQ-VAE to perform BERT-style self-supervised learning. Different from these approaches, our algorithm does not require training the quantizer which further simplifies the training process.

\section{Self-supervised Learning with Random-projection Quantizer}\label{method}

\begin{figure}[t]
    \includegraphics[width=\columnwidth]{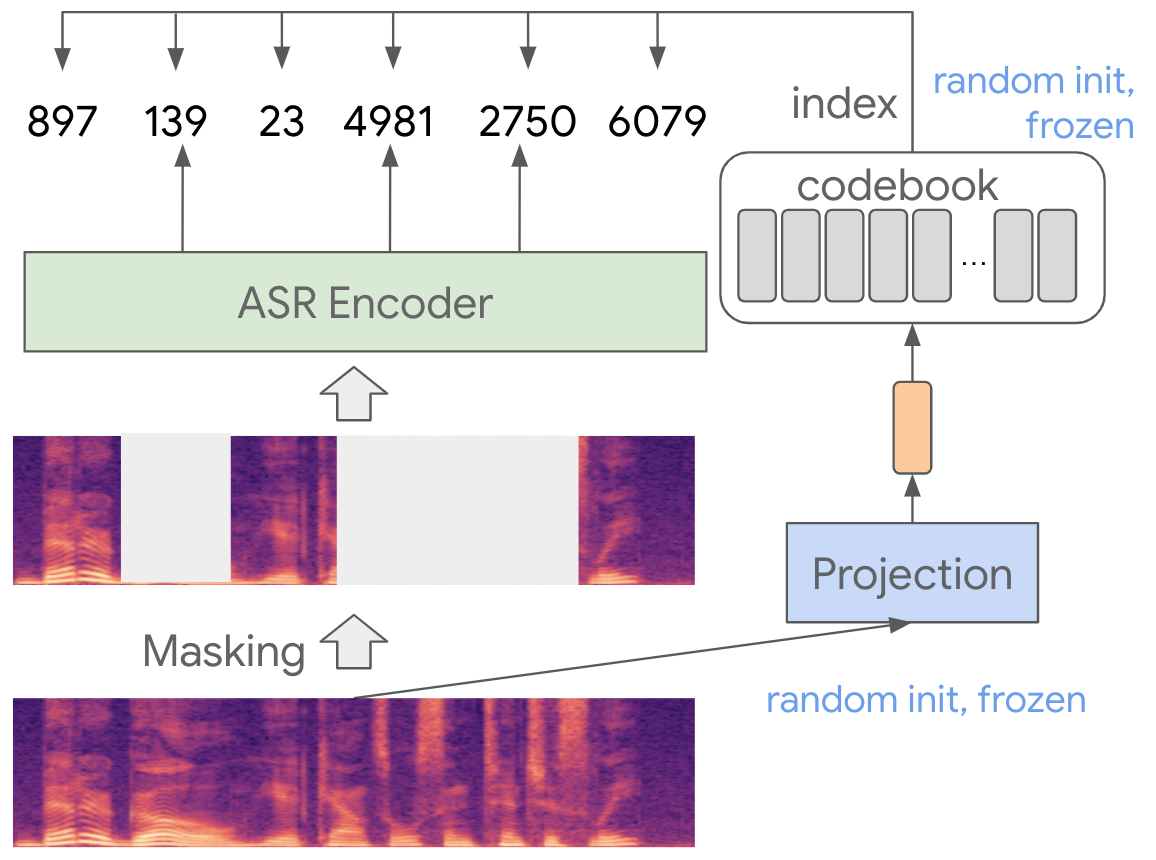}
    \caption{Overview of BEST-RQ. The approach applies random projections to project the input speech signals to a randomly initialized codebook, and map them to discrete labels through finding the nearest vector in the codebook. The pre-training objective is for the ASR encoder to take the masked input signals and predict the labels corresponding to the masked part provided by the random-projection quantizer.}
    \label{fig:randomproject}
\end{figure}

BEST-RQ applies a random-projection quantizer to map speech signals to discrete labels to enable BERT-style pre-training for ASR encoders. The quantizer randomly initializes a matrix and a codebook, and uses the matrix to project the input speech signals and the codebook to find the nearest vector where the index of the vector is the label. The pre-training process masks the speech signals and feeds them to the ASR encoder and trains the ASR encoder to predict labels of the masked part. Both the randomly initialized matrix and codebook are fixed during the pre-training process.  The input data is normalized to have $0$ mean and standard deviation of $1$. The normalization is critical for preventing the random projection to collapse to a small subset of codes. The framework is described in \cref{fig:randomproject}. After the pre-training process the resulting ASR encoder is adopted to fine-tune on downstream ASR tasks.

The approach applies masks directly on the speech signal, where the masking strategy samples at every frame whether to apply masks with a fixed probability. Each mask spans from the starting frame with a fixed length. The masked parts are replaced with a noise sampled from a normal distribution with $0$ mean and $0.1$ standard deviation.

\subsection{Random-projection Quantizer}

Given an input vector $x$ where $x$ is a $d$-dimensional vector computed from speech signals, the random-projection quantizer maps $x$ to discrete labels $y$ through
\begin{equation}
  y = \underset{i}{argmin}||norm_{l2}(c_{i}) - norm_{l2}(Ax)||,
\end{equation}
where $A$ denotes a randomly initialized $h\times d$ matrix and $C=\{c_{1},...,c_{n}\}$ is a set of randomly initialized $h$-dimensional vectors, $norm_{l2}()$ is a function that normalize the vector to have unit $l2$ norm. The projection matrix $A$ use Xavier initialization~\cite{glorot10} and the codebook $C$ use standard normal distribution for initialization, and the parameters are fixed during the pre-training process and therefore the quantizations are consistent during training.

\subsection{Pre-training}

The pre-training process adds a softmax layer on top of the ASR encoder to learn to predict the quantized speech labels. Since the random-projection quantizer is independent of the ASR encoder, the pre-training is flexible and can work with different architectures of the ASR encoder. We study the effectiveness of the algorithm on both non-streaming and streaming models, and in our experiments we use Conformer~\cite{gulati2020conformer} as the building block.

\subsubsection{Non-streaming Models}

Since the BERT-style pre-training is designed for the non-streaming models, training with this type of architecture is straightforward where the model uses both past and future context to learn to predict the quantized labels of the masked speech signals.

\subsubsection{Streaming Models}

In addition to the non-streaming models, streaming architecture also plays a critical role for the speech recognition tasks as many of the applications require transcribing speakers' utterances with low-latency \cite{Tara2020}. Streaming architecture however is less well-studied in the previous self-supervised learning work compared to the non-streaming architecture. Moreover, many of the previous self-supervised learning approaches specify a pre-training setup that takes both the previous and future context, making it a question of how one can generalize the approaches to streaming models. We proposed two pre-training algorithms that are compatible with the streaming architecture:

\textbf{Streaming pre-train}. As our algorithm does not require learning quantization and focuses only on training the ASR encoder, this approach largely benefits the streaming models. Pre-training for streaming models follows the same setup as non-streaming models but the ASR encoder now learns to predict the quantized labels of the masked part based only on the past context. 

\textbf{Non-Streaming pre-train}. Given that the neural network architecture like Transformer/Conformer allows switching from non-streaming to streaming behaviors by adding a mask for the future context within the same model, one can also perform pre-training with non-streaming setup for streaming models. Our algorithm provides benefits for streaming models with both non-streaming and streaming pre-training.

\subsection{Fine-tuning}

After the pre-training, the approach initializes the encoder of the downstream ASR from the pre-trained model, and fine-tunes on the supervised set. The softmax layer added on top of the encoder during the pre-training process is not used in fine-tuning. We focus on end-to-end models with RNN transducers~\cite{Graves12}, where the decoder uses LSTMs for the prediction network. On constructing the encoder, an additional projection layer is added on top of the pre-trained encoder to help it adapt to the downstream ASR task. The training process also updates the encoder during the supervised fine-tuning.

\subsection{Understanding the Effectiveness of the Random-projection Quantizer}

\begin{table*}[t]
  \caption{LibriSpeech results with non-streaming models. The LM used in our experiment is a Transfomer LM with model size $0.1$B.}\label{table:non-stream-librispeech}
  \centering
  \resizebox{0.9\width}{!}{%
  \begin{tabular}{lccccccccc}
    \toprule
    \bfseries Method & Size~(B) & \multicolumn{4}{c}{\bfseries No LM} & \multicolumn{4}{c}{\bfseries With LM} \\
    \cmidrule(r){3-6} \cmidrule(r){7-10}
    & & \bfseries dev & \bfseries dev-other & \bfseries test & \bfseries test-other
     & \bfseries dev & \bfseries dev-other & \bfseries test & \bfseries test-other \\
    \midrule
    wav2vec~2.0~\cite{wav2vec2} & 0.3 & 2.1 & 4.5 & 2.2 & 4.5 & 1.6 & 3.0 & 1.8 & 3.3 \\ 
    HuBERT~Large~\cite{hubert} & 0.3 & $-$ & $-$ & $-$ & $-$ & 1.5 & 3.0 & 1.9 & 3.3 \\
    HuBERT~X-Large~\cite{hubert} & 1.0 &  $-$ & $-$ & $-$ & $-$ &  1.5 & \textbf{2.5} & 1.8 & 2.9 \\
    w2v-Conformer~XL~\cite{ssllimit} & 0.6 & 1.7 & 3.5 & 1.7 & 3.5 & 1.6 & 3.2 & \bfseries 1.5 & 3.2 \\
    w2v-BERT~XL~\cite{w2vbert} & 0.6 & \textbf{1.5} & 2.9 & \bfseries 1.5 & \textbf{2.9} & \bfseries 1.4 & 2.8 & \bfseries 1.5 & 2.8 \\
    BEST-RQ (Ours) & 0.6 & \textbf{1.5} & \textbf{2.8} & 1.6 & \textbf{2.9} & \textbf{1.4} & 2.6 & \textbf{1.5} & \textbf{2.7} \\
    \bottomrule
  \end{tabular}
  }
\end{table*}

Our algorithm use a random-projection quantizer for self-supervised learning, and such a design raise two questions: how good is the resulting quantization quality with this quantizer and how much does the quantization quality affect the effectiveness of the self-supervised learning? We address these two questions through comparing our quantizer with VQ-VAEs. Using random-projections for quantizing speech signals shares some similarity as VQ-VAEs. The random projection performs dimension reduction for the speech signals while the random codebook provides an approximated discrete representation of the speech data distribution. VQ-VAEs also provide a discrete representation for the speech signals, but do so by learning a representation in the latent space that best preserves the speech data. Thus, comparing with VQ-VAEs gives us insight on the quantization quality of our quantizer and the effect of representation learning for self-supervised learning.

We demonstrate that the quantization quality of the random-projection quantizer is not ideal but yet effective for self-supervised learning by comparing it with VQ-VAE-based quantizations in \cref{sec:vq-vae}. We also show that the gap in the quantization quality is less an issue with the increase of the unsupervised data in \cref{sec:scaling}. The main objective of self-supervised learning for speech recognition is to train the model to learn contextual information. The random-projection quantizer preserve the distribution of the speech data, and in order for the model to learn to predict the quantized token based on unmasked signals, the model needs to learn to process the raw signals and infer the contextual information among speech data. Such a criterion allows the model to perform effective self-supervised learning with a random-projection quantizer.

\section{Experiments}\label{exp}

We perform self-supervised learning experiments on LibriSpeech with non-streaming and streaming models, and assess the approach on multilingual tasks with non-streaming models. We study the quantization quality of the random-projection quantizer by comparing it with the quantizer learned with VQ-VAEs. The implementation use Lingvo \cite{lingvo} library.

\subsection{LibriSpeech}

Following~\cite{ssllimit}, we conduct experiments on the LibriLight dataset~\cite{librilight} for pre-training, and fine-tune on the LibriSpeech training set which contains $960$ hours of data.  The input speech signals are $80$-dimensional log-mel filter bank coefficients, and each frame has the stride of $10$ms.  In the fine-tuning phase, the decoder has a vocab size $1024$ and uses a $1024$-token WordPiece model~\cite{wordpiece_schuster} for tokenizations that is constructed from the transcripts of the LibriSpeech training set.

\subsubsection{Non-streaming Models}

We use the same architectures reported in~\cite{ssllimit} for fair comparisons. The model has two convolution layers at the bottom which provide $4$ times temporal-dimension reduction for the input sequences. The rest of the layers are a stack of Conformer models. We explore $0.6$B model size which is extensively studied in the previous works. The model contains $24$ layers of Conformer models.

\textbf{Pre-train}. The pre-training uses mask length $400$ms with masking probability of $0.01$. The learning rate schedule uses a transformer learning rate schedule~\cite{transformer}. The training of the model uses Adam optimizer~\cite{adam} with $0.004$ peak learning rate and $25000$ warm-up steps. The batch size is $2048$. Since the encoder has $4$ times temporal-dimension reduction, the quantization with random projections stacks every $4$ frames for projections. The vocab size of the codebook is $8192$ and the dimension is $16$.

\textbf{Fine-tune}. The fine-tuning model also follow the same architecture as in~\cite{ssllimit} and use RNN Transducer (RNN-T)~\cite{Graves12} for decoder with $2$ layers of unidirectional LSTMs, where the hidden dimension of LSTMs are $1280$. The fine-tuning process use the Transformer learning rate schedule. Since the encoder is initialized from a pre-trained model, the fine-tuning process use a lower learning rate for the encoder than the decoder. The encoder use $0.0003$ peak learning rate and $5000$ warmup steps, while the decoder use $0.001$ peak learning rate and $1500$ warmup steps.

The results of pre-training with LibriLight and fine-tuning on LibriSpeech, along with comparisons with previous works, are shown in \cref{table:non-stream-librispeech}. Our results with LM use shallow fusion to incorporate the LM. The LM is a $0.1$B Transformer model trained on the LibriSpeech LM corpus, and the model has $8$ layers, $1024$ model dimension, and $4096$ feed-forward network dimension. By using the same architecture and similar optimization strategy as~\cite{ssllimit}, our approach shows similar WERs as previous best results on LibriSpeech both with and without LM.

\subsubsection{Streaming Models}

\begin{table*}[t]
  \caption{LibriSpeech results compared with previous works with the same streaming architecture, and use LibriLight set for pre-training and LibriSpeech $960$h set for fine-tuning. The relative latency (the lower the better) is the average difference of the word prediction time when comparing with the baseline Conformer $0.1$B model. Our algorithm outperforms wav2vec 2.0 and w2v-BERT on both WERs and latency.}\label{table:stream-librispeech}
  \vskip 0.1in
  \centering
  \begin{tabular}{lcccccc}
    \toprule
    \bfseries Method & Size~(B) & \bfseries dev & \bfseries dev-other & \bfseries test & \bfseries test-other & Relative latency (ms)\\
    \midrule
    Conformer $0.1$B & 0.1 & 4.1 & 10.3 & 4.5 & 9.8 & 0 \\
    Conformer $0.6$B & 0.6 & 3.9 & 9.8 & 4.4 & 9.4 &  15.3 \\
    \midrule
    \bfseries Non-Streaming pre-train \\
    wav2vec 2.0 & 0.6 & 2.6 & 7.3 & 3.0 & 7.2 & -10.1 \\
    w2v-BERT & 0.6 & 2.8 & 7.2 & 3.3 & 6.9 & -0.7 \\
    BEST-RQ (Ours) & 0.6 & \textbf{2.5} & \textbf{6.9} & \textbf{2.8} & \textbf{6.6} & -16.3 \\
    \midrule
    \bfseries Streaming pre-train \\
    wav2vec 2.0 & 0.6 & 2.7 & 8.0 & 2.9 & 7.9 & -130.6 \\
    w2v-BERT & 0.6 & 2.7 & 8.4 & 3.0 & 8.1 & -117.1 \\
    BEST-RQ (Ours) & 0.6 & \textbf{2.5} & \textbf{6.9} & \textbf{2.8} & \textbf{6.6} & \textbf{-130.9} \\
    \bottomrule
  \end{tabular}
\end{table*}

The architecture we use for the streaming experiments follows a similar design as previous work for building streaming ASRs~\cite{fastemit}. We scale the model size to be also $0.6$B to be consistent with the non-streaming experiments. The architecture has $3$ Conformer layers at the bottom, followed by a stacking layer with $2$ times temporal-dimension reduction and $20$ Conformer layers on top of the stacking layer. The Conformer has $1024$ hidden dimension for the self-attention layer and $4096$ for the feed-forward layers. The self-attention layer attends to the current and the previous $64$ frames, and the convolution has a kernel that covers the current and the past $3$ frames.

The training setup is mostly the same as the $0.6$B model in the non-streaming experiments, with some changes on the masking ratio for different pre-training approaches.

\textbf{Streaming pre-train}. The streaming pre-training uses the same setup as the original architecture, and the mask length is $300$ms and the masking probability is $0.02$. The random-projection quantizer stacks every $2$ frames for projections.

\textbf{Non-streaming pre-train}. The non-streaming pre-training extends the original architecture to have access for future context by having the convolution kernel within the Conformer layer to have access for the future $3$ frames. The self-attention is still limited to having access only for the previous context. We also explored having future context access for the self-attention, but this setup tends to be less stable. The masking length is $400$ms and the masking probability is $0.02$.

\textbf{Fine-tune}. The fine-tuning ASR model uses RNN-T for decoder with a layer of unidirectional LSTM, where the hidden dimension of the LSTM is $640$. The training setup is the same as the fine-tuning config for the $0.6$B model in the non-streaming experiments. When initializing from a non-streaming pre-trained model, the convolution only uses the kernel weight that accesses the previous context to keep the model streaming.

\textbf{Latency measurement}. A streaming model can learn to delay its prediction to access the future context and improve the prediction accuracy, and therefore it is critical to measure the latency of the streaming models to see whether the model maintains similar latency. This assessment helps us identify whether the underlying approach provides real improvement instead of trading off latency for prediction accuracy. Our latency comparison process first calculates the starting time and ending time of every word for each hypothesis generated by the two models. The comparison then aligns the hypotheses from the two models, finds the matching words and calculates the difference of their starting and ending time. The relative latency measurement is the average word timing difference of all matched words between the two models among all utterances. Specifically the relative latency is calculated with
\begin{equation}
  \sum_{i,j}\frac{s'_{ij} - s_{ij} + e'_{ij} - e_{ij}}{2N},
\end{equation}
where $i$ denotes the index of the matched words between the two hypotheses, $j$ is the utterance index, $s_{ij}$ and $e_{ij}$ correspond to the starting and ending time of the word from the baseline model, $s'_{ij}$ and $e'_{ij}$ correspond to the starting and ending time of the word from the compared model, and $N$ is the total number of matched word among all utterances. A negative relative latency means the compared model has lower latency than the baseline model.

The word-error-rates and the relative latency are shown in~\cref{table:stream-librispeech}. In this comparison experiment both wav2vec 2.0 and w2v-BERT use the same architecture, same masking and training setup as BEST-RQ. Using the conventional masking setup for wav2vec 2.0 and w2v-BERT gives worse performance. Since there is no convolution layers at the bottom, the contrastive learning use speech signals as targets. The w2v-BERT model use $12$ layers for the contrastive module and $12$ layers for the masked prediction module, to be consistent with the non-streaming setup~\cite{w2vbert}. Our algorithm outperforms wav2vec 2.0 and w2v-BERT for both streaming and non-streaming pre-training. In particular our algorithm performs well with both pre-training, while wav2vec 2.0 and w2v-BERT favors more with non-streaming pre-training. This is likely due to the fact that the representation learning of both approaches is more compatible with non-streaming architectures. Increasing the model size from $0.1$B to $0.6$B results a slight increase in latency, but models trained with self-supervised learning algorithms has lower latency with streaming pre-training giving the most significant latency reduction. This indicates that the self-supervised learning preserve the low-latency property while providing quality gain.

\vspace{-0.2cm}
\subsection{Multilingual Tasks}

We present mutlilingual results in this section. We use the same model setup as the LibriSpeech non-streaming experiment for these tasks.

\subsubsection{Data}

\paragraph{Multilingual LibriSpeech(MLS-10hrs)}
The Multilingual LibriSpeech dataset~\cite{pratap2020mls} is a large corpus derived from read audiobooks of Librivox and consists of 8 languages: \textit{Dutch (du), English (en), French (fr), German (de), Italian (it), Polish (pl), Portuguese (pt), Spanish (es)}. The latest version of this corpus contains around 50k hours including 44k hours in English. We use the official 10 hours split of training data to evaluate few-shot learning capabilities.

\paragraph{Multilingual Voice Search (VS-1000hrs)}
Our high resource finetune datasets is multilingual Voice Search dataset \cite{li2021scaling}. We sample random $1000$ hour subsets (VS-1000h) across 15 languages, including English (US), English (IN), Spanish (US), Portuguese (BR), Spanish (ES), Arabic (GULF), Arabic (EG), Hindi (IN), Marathi (IN), Bengali (BD), Chinese (TW), Russian (RU), Turkish (TR), Hungarian (HU), and Malay (MY). The test set for each language contains around 3--19K utterances. 

\paragraph{\xlsrp unsupervised data ({\xlsrp}-U)}
Our public unlabeled speech data follows the pre-training data used for \xlsrp~\citep{babu2021xls} with one major difference: we do not use any data from VoxLingua-107 due to license constraint. In total, we utilize approximately $429k$ hours of unlabeled speech data in $51$\footnote{Counting languages with more than $1$ hour of speech data.} languages.  As a consequence our model is pre-trained on speech from $51$ languages as compared to $128$ for \xlsrp, and our pre-training set is smaller by $6.6k$ hours. We use this pretrain data on MLS-10hrs to compare with published results.

\paragraph{Youtube unsupervised data (YT-U)}
Following \cite{bigssl}, we collected a multilingual Youtube dataset for pretraining. For each language we prepare an unlabeled YouTube dataset segmented using voice activation detection (VAD \cite{zazo2016feature}). The number of hours per languages are: English (800k hrs), Spanish (800k hrs), Marathi (600k hrs), Portuguese (800k hrs), Russian (800k), Arabic (800k), Hindi (800k), Chinese (800k), Malay (250k), Turkish (800k), Bengali (800k), Hugarian (300k). In practice, we found this data performs much better than {\xlsrp}-U on VS-1000hrs. Thus, we use this pretrain data on VS-1000hrs to compare the performance of different pretrain methods.

\begin{table*}[t]
\caption{Test set WER (\%) comparisons on the MLS full and 10hrs set.}
\centering
\begin{tabular}{lccccccccc}
\toprule 
\multirow{2}{*}{\tbh{Exp.}} & \multicolumn{8}{c}{\tbh{Languages}} &  \multirow{2}{*}{\tbh{Avg.}} \\
\cmidrule(lr){2-9}
~ & {\bf en }& {\bf de }& {\bf nl }& {\bf fr }& {\bf es }& {\bf it }& {\bf pt} & {\bf pl} & ~\\
\midrule
\multicolumn{10}{l}{\it \bf MLS-full} \\
\midrule
wav2vec 2.0 from XLSR-53 \cite{conneau2020unsupervised} & - & 7.0 & 10.8 & 7.6 & 6.3 & 10.4 & 14.7 & 17.2 & 10.6 \\
w2v-BERT from JUST  \cite{bai2021joint} & 6.6 & 4.3 & 9.9 & 5.0 & 3.8 & 9.1 & 14.6 & 8.1 & 7.8 \\
JUST \cite{bai2021joint} (co-train) & 6.5 & 4.1 & 9.5 & 5.2 & 3.7 & 8.8 & 8.0 & 6.6 & 6.5 \\
\midrule
w2v-BERT (0.6B) & 5.5 & 4.3	& 10.9 & 5.6	& 4.5	& 10.1 & 13.4 & 11.2 & 8.2\\
BEST-RQ (Ours, 0.6B) & 6.8 & 4.1 & 9.7 & 5.0 & 4.9 & 7.4 & 9.4 & 5.2 & 6.6\\
\midrule
\midrule
\multicolumn{10}{l}{\it \bf MLS-10hrs} \\
\midrule
XLSR-53 \cite{conneau2020unsupervised}& 14.6 & 8.4 & 12.8 & 12.5 & 8.9 & 13.4 & 18.2 & 21.2  & 13.8 \\
XLS-R(0.3B) \cite{babu2021xls} & 15.9 & 9.0 & 13.5 & 12.4 & 8.1 & 13.1 & 17.0 & 13.9 & 12.8 \\
XLS-R(1B) \cite{babu2021xls} & 12.9 & 7.4 & 11.6 & 10.2 & 7.1 & 12.0 & 15.8 & 10.5 & 10.9 \\
XLS-R(2B) \cite{babu2021xls} & 14.0 & 7.6 & 11.8 & 10.0 & 6.9 & 12.1 & 15.6 & 9.8 & 11.0 \\
\midrule
w2v-BERT (0.6B) & 12.7 & 7.0 & 12.6 & 8.9 & 5.9 & 10.3 & 14.6 & 6.9 & 9.9\\
BEST-RQ (Ours, 0.6B) & 12.8 & 7.4 & 12.7 & 9.6 & 5.4 & 9.9 & 12.1 & 7.1 & 	9.6\\
\bottomrule
\end{tabular}
\label{tbl:mls}
\end{table*}

\subsubsection{Results on MLS-10hrs}

We conduct our multilingual low resource finetune experiments on MLS-10hrs. We use {\xlsrp}-U as pretraining data and finetune it on MLS-10hrs. As shown in Table~\ref{tbl:mls}, our baseline w2v-BERT already outperform previous strong model from XLS-R(2B) \cite{babu2021xls}. The average WER further bring down by 3\% relative by using the proposed BEST-RQ. This demonstrate a simple random-projection quantizer is also effective for multilingual pretraining. We also report finetune results on the MLS full supervised data. Interestingly, with more finetune data, BEST-RQ perform even better than w2v-BERT, especially for pt and pl. Our results also comparable with previously state-of-the-art results in \cite{bai2021joint} which conduct joint training for multilingual ASR.

While fine-tuning with MLS-full and MLS-10hrs both exhibit improvement compared to existing approaches, fine-tuning with MLS-full provides more relative improvement. This likely implies that pre-training with random-projection quantizers is more effective when there is more fine-tuning data.

\subsubsection{Results on Voice Search}
\begin{table}[t]
\caption{Test set WER (\%) comparisons using YT-U for pretrain and VS-1000hrs for finetune, across 15 languages.}
\centering
\begin{tabular}{lc}
\toprule 
\multirow{1}{*}{\tbh{Exp.}} & \tbh{Avg.} on 15 langs (VS)\\
\midrule
Baseline (0.6B) & 12.6 \\
wav2vec 2.0 (0.6B) & 12.0\\
w2v-bert (0.6B) & 11.5\\
BEST-RQ (Ours) (0.6B) & \tbh{10.9}\\
\bottomrule
\end{tabular}
\label{tbl:vs}
\end{table}

To understand how the proposed model work for high resource (1000hrs per language), we pretrain our model on YT-U and finetune it on VS-1000hrs. We can see with more finetune data, the relative improvement is smaller compared with no pretrain baseline. However, our proposed BEST-RQ consistently outperform w2v-BERT and wav2vec 2.0 by 9\% and 5\% relatively. Compare to w2v-BERT, our proposed method outperform on all the languages. Among the 15 languages, English, Portuguese, Russian and Turkish, are improved more than 10\%, relatively. Indic languages (Hindi, Marathi and English (IN)) are only slightly improved, all smaller than 3\% relatively. 

\subsection{Analyzing Quantization Quality}\label{sec:vq-vae}

\begin{table*}[t] 

  \caption{Quantizer quality's impact on ASR tasks. Although the Transformer-based quantizer gets much better performance when used as input directly, the random-projection quantizer is equally effective for self-supervised learning. The model used in the direct ASR task has size $25$M. The self-supervised learning tasks use the same setup as the LibriSpeech non-streaming experiment, which use LibriLight for pre-training and LibriSpeech for fine-tuning and has $0.6$B model size.}\label{table:vq-vae}
  \centering
  \begin{tabular}{lccccccccc}
    \toprule
    Configuration & Quantizer size (M) & \multicolumn{4}{c}{Direct ASR WER} & \multicolumn{4}{c}{Pretrain-finetune WER} \\
    \cmidrule(r){3-6} \cmidrule(r){7-10} & & dev & dev-other & test & test-other & dev & dev-other & test & test-other \\
    \midrule
    Random quantizer & 1 & 58.8 & 78.8 & 57.9 & 72.8 & 1.5 & \textbf{2.8} & \textbf{1.6} & \textbf{2.9} \\ 
    Projection VQ-VAE & 1 & 61.4 & 74.8 & 60.9 & 75.2 & 1.5 & \textbf{2.8} & \textbf{1.6} & \textbf{2.9} \\
    Transformer VQ-VAE & 10 & \textbf{17.8} & \textbf{35.8} & \textbf{17.6} & \textbf{36.1} & \textbf{1.4} & 2.9 & \textbf{1.6} & 3.1\\
    \bottomrule
  \end{tabular}
\end{table*}

As our self-supervised learning algorithm eliminates the requirement of representation learning through applying a random-projection quantizer, it is crucial to understand the representation quality of this quantizer and how the quality of the quantization affect the self-supervised learning. We analyze the quality of quantizers by training ASR models feeding labels generated by quantizing utterances as input. The performance of the resulting ASR provides us insights on the quality of the quantizer. The ASR model embeds quantized labels and feeds the embedding to a stack of Conformer layers, followed by a CTC decoder. $16$ Conformer layer has feature dim $256$, local self attention with $8$ heads and $128$ context length and kernel size $5$ for lightweight convolution, in total the model size is $25$M. We study the effect of representation learning through comparing with quantizers trained with the VQ-VAE. We compare $3$ types of quantizers: a) a random-projection quantizer b) a quantizer trained with VQ-VAE where the encoder has the same architecture as the random-projection quantizer and the decoder contains only a projection layer c) a trained VQ-VAE whose encoder/decoder are Transformer models.  For trained quantizers, we train on the whole LibriSpeech $960$ hours audio-only data, with a constant learning rate of 1e-4 and train for $400$k steps with batch size $256$. For all quantizers, the input frames are stacked with 3 frames on each's left, resulting in 4x input length reduction. We also use the quantizers for self-supervised learning with the LibriSpeech $0.6$B non-streaming setup to compare their performance.

\cref{table:vq-vae} shows the WER on LibriSpeech $960$h. Both the random-projection quantizer and the projection-based VQ-VAE quantizer lead to poor ASR performance, while the Transformer-based VQ-VAE quantizer provides a significantly better performance. This implies that the Transformer-based VQ-VAE quantizer learns a better representation. On the other hand, when using these quantizers for the purpose of self-supervised learning, all quantizers lead to similar WERs. This indicates that the quantizer quality does not translate to self-supervised learning quality.

\subsection{Analyzing the Effect of Pre-training Data Size}\label{sec:scaling}

\begin{figure}[t]
    \includegraphics[width=\columnwidth]{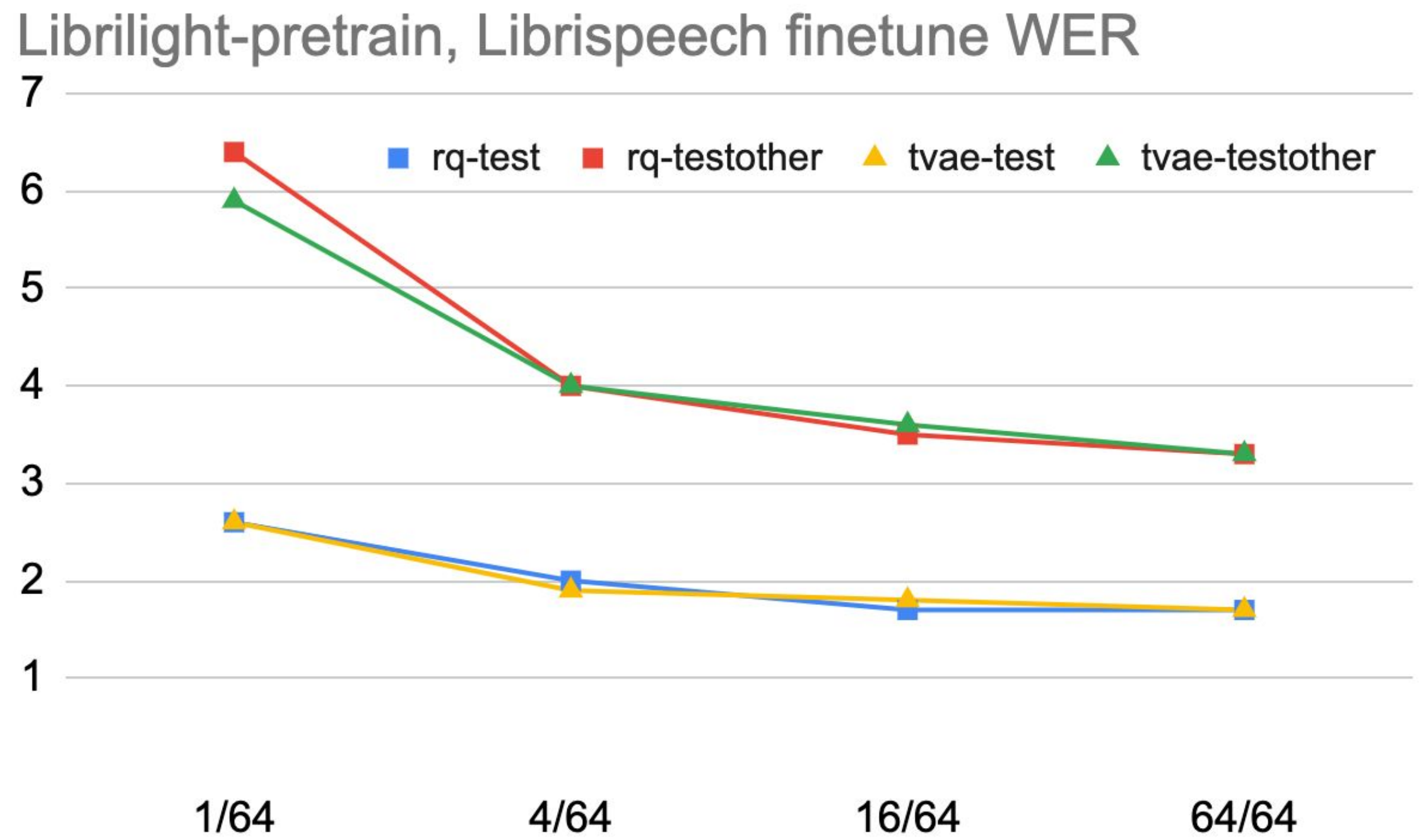}
    \caption{Comparing the self-supervised learning quality of the random-projection quantizer (rq) and the Transformer-based VQ-VAE quantizer (tvae) with different pre-training data size. Starting from low amount of pre-train data, the random-projection quantizer is behind the trained Transformer VQ-VAE quantizer. As the amount of pre-train data increases, the random-projection quantizer catches up.}
    \label{fig:datascaling}
    \vspace{-0.2cm}
\end{figure}

One potential explanation for the above observation, that a sub-optimal quantization can work well for self-supervised learning, is that the self-supervised learning algorithm can learn to mitigate the quality gap given sufficient amounts of pre-training data. We investigate whether a quantizer with a better quantization quality performs better when the amount of the pre-training data is limited, and whether increasing the amount of the pre-training data alleviate the discrepancy when compared to a random-projection quantizer. In this study, we compare self-supervised learning quality between a random-projection quantizer (rq) and a trained transformer-based VQ-VAE quantizer (tvae) with different pre-training data sizes. The random quantizer is un-trained, and 4 Transformer VQ-VAE quantizers are trained with  $\{1/64, 4/64, 16/64, 64/64\}$ LibriLight data, respectively. Then 4 identical random-projection quantizers and the above 4 transformer VAE quantizers are pre-trained separately with the same distinct percentages of LibriLight data as above for 100k steps with global batch size 2048. The pre-trained models fine-tune on LibriSpeech 960h. The result in \cref{fig:datascaling} shows that a quantizer with better representation quality (Transformer-based VQ-VAE) performs better when pre-training data is limited, but the gap disappears as the pre-training data increase.

\section{Conclusions and Discussions}

We proposed BEST-RQ to perform self-supervised learning for speech recognition models. BEST-RQ uses a random-projection quantizer to quantize speech signals to discrete labels. The pre-training process masks the speech signals and trains the model to predict labels corresponding to the masked parts. This approach shows similar WERs as the existing state-of-the-art results on LibriSpeech with non-streaming models, and outperform wav2vec 2.0 and w2v-BERT on LibriSpeech with streaming models and on multilingual tasks with non-streaming models. Further analysis showed that despite the fact that the random-projection quantizer provides a poorer representation compared to a trained VQ-VAE quantizer, it is effective for the purpose of self-supervised learning.

Our algorithm untangle the quantizer from the speech recognition model and also eliminates the requirement of representation learning. This simpler framework makes it easier to find a good recipe for the target task. The improvement on streaming models shows that the separation of the quantizer from the model makes the algorithm more effective for architectures that can be less effective for representation learning. The improvement on multilingual tasks shows that complicated tasks can benefit more from a simpler framework where finding a good recipe becomes more challenging. The quantization quality analysis implies that representation learning is not necessarily critical for self-supervised learning.

\tbh{Codebook utilization.} One of the most critical factors for pre-training quality is the percentage of the codebook that is used during training. In particular, at each training step a higher percentage of the codebook being used in each batch correlates strongly with a good pre-training quality. When the distribution of the codebook utilization is skewed toward a smaller subset of codes, this usually makes the pre-training task easier and provides less effective pre-training. The $l2$ normalizations on the projected vector and the codebook are critical for providing more uniform codebook utilization. On the other hand, using randomly initialized codebook and projection matrix can introduce different codebook utilizations with different random seeds, which impact the pre-training quality across different runs with same experiment configurations. This variance impacts quality more when training with smaller pre-training and fine-tuning datasets. How to reduce this reproducibility issue caused by random initialization is an important next step for improving random-projection quantizations.

\tbh{Hyperparameters.} The pre-training quality is not very sensitive to the codebook vocab size and the codebook dimension, and is more sensitive to the masking probability and the mask length. The role of the projection layer in the random-projection quantizer is to allow using different codebook dimensions, and one can achieve similar results without the projection and set the codebook dimension to be the same as the input dimension. Due to the variance coming from the random initialization, the impact of a hyperparameter usually requires multiple runs of experiments to verify the result.

\tbh{Longer convergence time for non-streaming models.} One observation we have is that the algorithm takes more steps to converge with non-streaming models. We still observe improvement compared to wav2vec 2.0 and w2v-BERT at the same training step on multilingual tasks, though the final convergence usually takes $50\%$ more steps. On the other hand, our training setup follows~\cite{ssllimit}, and it is unclear to us whether further hyperparameter tuning can help the model to converge faster. We did not observe the longer convergence property with streaming models.

\tbh{Initialization.} The quantizer uses random initialization and does not update the parameters, and therefore the initialization algorithm can play an important role on the results. In this paper we showed results with Xavier initialization for the projection matrix and the standard normal distribution for the codebook, and further comparisons on different initialization algorithms can be conduct in the future work.

\section{Acknowledgements}

We thank Wei Han and Johan Schalkwyk for helpful discussions, and Rohit Prabhavalkar, Izhak Shafran, and Hagen Soltau for insightful feedback. We also want to thank Bo Li for the help on multilingual tasks.

\bibliography{main}
\bibliographystyle{main}

\end{document}